\documentclass[11pt]{article}

\usepackage[final,nonatbib]{wiml_abstract}

\DeclareMathSizes{10}{8}{6}{5}
\usepackage[
    backend=biber,
    citestyle=authoryear, 
    maxcitenames=1,
    sorting=none,
    natbib=true, 
    ]{biblatex}

\usepackage{scrhack} 
\usepackage{algorithm}
\usepackage[noend]{algpseudocode}

\algrenewcommand{\algorithmiccomment}[1]{\hfill {\small \textcolor{darkgray}{$\vartriangleright$ #1}}}
\algrenewcommand\algorithmicindent{2em}
\algrenewcommand\alglinenumber[1]{\small {\textcolor{darkgray}{#1}}}

\makeatletter
\expandafter\patchcmd\csname\string\algorithmic\endcsname{\itemsep\z@}{\itemsep=0.25ex}{}{}
\newcommand\fs@booktabsruled{%
    \def\@fs@cfont{\bfseries\strut}\let\@fs@capt\floatc@ruled
    \def\@fs@pre{\hrule height\heavyrulewidth depth0pt \kern\belowrulesep}%
    \def\@fs@mid{\kern\aboverulesep\hrule height\lightrulewidth\kern\belowrulesep}%
    \def\@fs@post{\kern\aboverulesep\hrule height\heavyrulewidth\relax}%
    \let\@fs@iftopcapt\iftrue
}
\makeatother
\floatstyle{booktabsruled}
\restylefloat{algorithm}
\usepackage{xcolor}

\definecolor{MPLblue}{HTML}{1f77b4}
\definecolor{MPLorange}{HTML}{ff7f0e}
\definecolor{MPLgreen}{HTML}{2ca02c}
\definecolor{MPLred}{HTML}{d62728}
\definecolor{MPLpurple}{HTML}{9467bd}

\definecolor{SNSblue}{rgb}{0.1216, 0.4666, 0.7059}
\definecolor{SNSorange}{rgb}{1.0, 0.4980, 0.0549}
\definecolor{SNSgreen}{rgb}{0.1725, 0.6274, 0.1725}
\definecolor{SNSred}{rgb}{0.84, 0.15, 0.16}
\definecolor{SNSpurple}{rgb}{0.58, 0.40, 0.74}

\definecolor{SNSblue_shaded}{HTML}{8ebad9}
\definecolor{SNSorange_shaded}{HTML}{ffcea3}
\definecolor{SNSgreen_shaded}{HTML}{cae7ca}
\definecolor{SNSred_shaded}{HTML}{ea9293}

\definecolor{TUgray}{RGB}{185,184,188}
\definecolor{TUdarkgray}{RGB}{50,65,75}
\definecolor{TUgold}{RGB}{180,160,105}
\definecolor{TUdarkblue}{RGB}{65,90,140}
\definecolor{TUblue}{RGB}{0,105,170}
\definecolor{TUlightblue}{RGB}{80,170,200}
\definecolor{TUlightgreen}{RGB}{130,185,160}
\definecolor{TUgreen}{RGB}{125,165,75}
\definecolor{TUdarkgreen}{RGB}{50,110,30}
\definecolor{TUred}{RGB}{165,30,55}
\definecolor{TUlightred}{RGB}{200,80,60}
\definecolor{TUorange}{RGB}{210,150,0}
\definecolor{TUpurple}{RGB}{175,110,150}
\definecolor{TUocre}{RGB}{200,80,60}
\definecolor{TUviolet}{RGB}{175,110,150}
\definecolor{TUmauve}{RGB}{180,160,150}
\definecolor{TUbeige}{RGB}{215,180,105}
\definecolor{TUbrown}{RGB}{145,105,70}

\definecolor{CUblue}{RGB}{29,79,145}
\definecolor{CUlightblue}{RGB}{185,217,235}
\definecolor{CUdarkblue}{RGB}{0,48,135}
\definecolor{CUgray}{RGB}{117,120,123}
\definecolor{CUlightgray}{RGB}{208,208,206}
\definecolor{CUdarkgray}{RGB}{83,86,90}
\definecolor{CUmagenta}{RGB}{174,37,115}
\definecolor{CUyellow}{RGB}{255,152,0}
\definecolor{CUorange}{RGB}{185,71,0}
\definecolor{CUred}{RGB}{166,10,61}
\definecolor{CUgreen}{RGB}{118,136,29}

\colorlet{maincolor}{CUblue}		        
\colorlet{secondcolor}{CUblue}		        

\colorlet{lightgraycolor}{CUlightgray}  
\colorlet{darkgraycolor}{TUdarkgray}	

\colorlet{alertcolor}{CUred}
\colorlet{alertlightcolor}{CUred!50!white}

\makeatletter
\@ifclassloaded{beamer}{
    \colorlet{citecolor}{CUblue}
    \colorlet{citelightcolor}{CUlightblue}
    \colorlet{linkcolor}{CUblue}
    \colorlet{linklightcolor}{CUlightblue}
    \colorlet{urlcolor}{CUblue}
    \colorlet{urllightcolor}{CUlightblue}
}{
    \colorlet{citecolor}{MPLblue}
    \colorlet{linkcolor}{black}
    \colorlet{urlcolor}{MPLblue}
}

\makeatother

\usepackage{enumitem}

\usepackage{caption}
\usepackage[labelformat=simple]{subcaption}
\usepackage{wrapfig}
\usepackage{needspace}

\usepackage{graphicx}
\usepackage{pgfplots}
\pgfplotsset{compat=1.15}
\usepackage{adjustbox}
\usepackage{tcolorbox}
\usepackage{tikz}

\newcommand{\colordot}[2][0.75ex]{\tikz[baseline=-0.66ex]\draw[#2,fill=#2,radius=#1,fill opacity=0.4] (0,0) circle ;}%

\usepackage{amsmath}
\usepackage{amssymb}
\usepackage{mathtools} 
\usepackage{bm}
\usepackage{etoolbox}

\allowdisplaybreaks    

\DeclarePairedDelimiterX{\set}[1]\{\}{%

#1}

\DeclarePairedDelimiterXPP\linspan[1]{\operatorname{span}}{(}{)}{}{#1}
\DeclarePairedDelimiterXPP\im[1]{\operatorname{im}}{(}{)}{}{#1}
\let\ker\relax 
\DeclarePairedDelimiterXPP\ker[1]{\operatorname{ker}}{(}{)}{}{#1}
\DeclarePairedDelimiterXPP\rank[1]{\operatorname{rank}}{(}{)}{}{#1}
\DeclarePairedDelimiterXPP\trace[1]{\operatorname{tr}}{(}{)}{}{#1}
\let\det\relax
\DeclarePairedDelimiterXPP\det[1]{\operatorname{det}}{(}{)}{}{#1}

\DeclareSymbolFont{stmry}{U}{stmry}{m}{n}
\DeclareMathSymbol\obar\mathrel{stmry}{"3A}
\DeclareMathSymbol\otimes\mathrel{stmry}{"0F}
\DeclareMathSymbol\ominus\mathrel{stmry}{"17}
\makeatletter
\newcommand{\superimpose}[2]{
  {\ooalign{$#1\@firstoftwo#2$\cr\hfil$#1\@secondoftwo#2$\hfil\cr}}}
\makeatother

\makeatother

\DeclarePairedDelimiterX\abs[1]{\lvert}{\rvert}{
  \ifblank{#1}{\:\cdot\:}{#1}
}
\DeclarePairedDelimiterX\norm[1]{\lVert}{\rVert}{
  \ifblank{#1}{\:\cdot\:}{#1}
}
\DeclarePairedDelimiterX\innerprod[2]{\langle}{\rangle}{
  \ifblank{#1}{\:\cdot\:}{#1},\ifblank{#2}{\:\cdot\:}{#2}}

\DeclareDocumentCommand\partialderivative{ s o m g g d() }
{ 
  \IfBooleanTF{#1}
  {\let\fractype\flatfrac}
  {\let\fractype\frac}
  \IfNoValueTF{#4}
  {
    \IfNoValueTF{#6}
    {\fractype{\partial \IfNoValueTF{#2}{}{^{#2}}}{\partial #3\IfNoValueTF{#2}{}{^{#2}}}}
    {\fractype{\partial \IfNoValueTF{#2}{}{^{#2}}}{\partial #3\IfNoValueTF{#2}{}{^{#2}}} \argopen(#6\argclose)}
  }
  {
    \IfNoValueTF{#5}
    {\fractype{\partial \IfNoValueTF{#2}{}{^{#2}} #3}{\partial #4\IfNoValueTF{#2}{}{^{#2}}}}
    {\fractype{\partial^2 #3}{\partial #4 \partial #5}}
  }
}

\NewDocumentCommand{\jac}{o m m o}{%
  \operatorname{D}\IfNoValueF{#1}{_{#1}}\mathopen{}%
  #2\mathopen{}\left( #3 \right)\mathclose{}%
  \IfNoValueF{#4}{%
    \IfNoValueTF{#1}{%
      |_{#3}
    }{%
      |_{#1=#4}
    }
  }
}

\NewDocumentCommand{\grad}{o m m o}{%
  \nabla\IfNoValueF{#1}{_{#1}}\mathopen{}%
  #2\mathopen{}\left( #3 \right)\mathclose{}%
  \IfNoValueF{#4}{%
    \IfNoValueTF{#1}{%
      |_{#3}
    }{%
      |_{#1=#4}
    }
  }
}

\NewDocumentCommand{\natgrad}{o m m o}{%
  \tilde{\nabla}\IfNoValueF{#1}{_{#1}}\mathopen{}%
  #2\mathopen{}\left( #3 \right)\mathclose{}%
  \IfNoValueF{#4}{%
    \IfNoValueTF{#1}{%
      |_{#3}
    }{%
      |_{#1=#4}
    }
  }
}

\NewDocumentCommand{\hessian}{o m m o}{%
  \nabla^2\IfNoValueF{#1}{_{#1}}\mathopen{}%
  #2\mathopen{}\left( #3 \right)\mathclose{}%
  \IfNoValueF{#4}{%
    \IfNoValueTF{#1}{%
      |_{#3}
    }{%
      |_{#1=#4}
    }
  }
}

\makeatletter
\newcommand*{\@probsymbol}{\mathrm{P}}
\newcommand*{\@given}[1]{%
  \nonscript\:#1\vert
  \allowbreak
  \nonscript\:
  \mathopen{}}

\providecommand*{\given}{}
\DeclarePairedDelimiterXPP{\@prob}[1]{\@probsymbol}{(}{)}{}{%
  \renewcommand*{\given}{\@given{\delimsize}}%
  #1}
\newcommand*{\prob}[1]{\ifblank{#1}{\@probsymbol}{\@prob*{#1}}}
\makeatother

\NewDocumentCommand{\expval}{o o m}{%
  \operatorname{\mathbb{E}}\IfNoValueF{#1}{_{#1\IfNoValueF{#2}{\sim #2}}}\mathopen{}\left( #3 \right)
}
\NewDocumentCommand{\cov}{o o m}{%
  \operatorname{Cov}\IfNoValueF{#1}{_{#1\IfNoValueF{#2}{\sim #2}}}\mathopen{}\left( #3 \right)
}
\NewDocumentCommand{\var}{o o m}{%
  \operatorname{Var}\IfNoValueF{#1}{_{#1\IfNoValueF{#2}{\sim #2}}}\mathopen{}\left( #3 \right)
}

\NewDocumentCommand{\entropy}{o o m}{%
\operatorname{H}\IfNoValueF{#1}{_{#1\IfNoValueF{#2}{\sim #2}}}\mathopen{}\left( #3 \right)
}
\NewDocumentCommand{\infogain}{o o m}{%
\operatorname{IG}\IfNoValueF{#1}{_{#1\IfNoValueF{#2}{\sim #2}}}\mathopen{}\left( #3 \right)
}

\NewDocumentCommand{\dw}{O{2} m m}{%
  \operatorname{W}\IfNoValueF{#1}{_{#1}}\mathopen{}\left( \ifblank{#2}{\:\cdot\:}{#2},\ifblank{#3}{\:\cdot\:}{#3} \right)
}

\DeclarePairedDelimiterXPP\bigO[1]{\mathcal{O}}{(}{)}{}{#1}
\DeclarePairedDelimiterXPP\smallo[1]{o}{(}{)}{}{#1}
\DeclarePairedDelimiterXPP\bigOmega[1]{\Omega}{(}{)}{}{#1}
\DeclarePairedDelimiterXPP\smallomega[1]{\omega}{(}{)}{}{#1}
\DeclarePairedDelimiterXPP\bigTheta[1]{\Theta}{(}{)}{}{#1}

\makeatletter
\@ifclassloaded{beamer}{
  \usepackage{pdfpages}

  \hypersetup{
    pdfkeywords={SP-Right}
  }
}
\makeatother

\usepackage{titletoc}

\makeatletter
\@ifclassloaded{beamer}{}{
    \usepackage{tocloft}
    \setlength\cftbeforesecskip{0em}
    \setlength\cftbeforesubsecskip{-0.5em}
    \setlength\cftbeforesubsubsecskip{-0.5em}
    
}
\makeatother

\usepackage[numbered,open=true]{bookmark}

\usepackage{booktabs}
\usepackage{longtable}
\usepackage{multirow}
\usepackage{rotating}
\usepackage{siunitx}
\usepackage[normalem]{ulem}
\usepackage{amsthm}
\usepackage{thmtools}
\usepackage{thm-restate}

\declaretheoremstyle[
    headfont=\normalfont\bfseries,
    notefont=\normalfont,
    bodyfont=\normalfont,
    headpunct={},
    postheadspace=\newline,
]{definitionstyle}

\declaretheoremstyle[
    headfont=\normalfont\bfseries,
    notefont=\normalfont,
    bodyfont=\normalfont\itshape,
    headpunct={},
    postheadspace=\newline,
]{lemmastyle}

\declaretheoremstyle[
    headfont=\normalfont\bfseries,
    notefont=\normalfont,
    bodyfont=\normalfont\itshape,
    headpunct={},
    postheadspace=\newline,
]{theoremstyle}

\declaretheoremstyle[
    headfont=\normalfont\bfseries,
    notefont=\normalfont,
    bodyfont=\normalfont,
    headpunct={},
    postheadspace=\newline,
]{remarkstyle}

\makeatletter
\@ifclassloaded{beamer}{
    
}{

}
\makeatother

\usepackage{hyperref}
\hypersetup{
    unicode,                    
    pdfborder       = {0 0 0},  
    bookmarksdepth  = subsection,
    bookmarksopen   = true,     
    linktoc         = all,      
    breaklinks      = true,
    colorlinks      = true,
    linkcolor       = linkcolor,
    citecolor       = citecolor,
    urlcolor        = urlcolor,
}

\usepackage[capitalise,nameinlink,sort&compress,noabbrev]{cleveref}
\crefname{subfigure}{Fig.}{Figs.}
\Crefname{subfigure}{Fig.}{Figs.}
\crefname{figure}{Fig.}{Figs.}
\Crefname{figure}{Fig.}{Figs.}

\hypersetup{
    citecolor = black,
}

\definecolor{myorange}{RGB}{255,165,0}
\definecolor{myblue}{RGB}{0,0,128}
\definecolor{mygreen}{RGB}{0,128,0}
\definecolor{myred}{RGB}{255,10,10}
\definecolor{myviolet}{RGB}{138,43,226}

\definecolor{observed}{RGB}{170,170,170} 
\definecolor{inference}{RGB}{255,91,89}  
\definecolor{gptfouro}{HTML}{08306B}
\definecolor{claude}{HTML}{66C2A5} 
\definecolor{gemini}{HTML}{A6D854}
\definecolor{humans}{HTML}{E7298A}
\definecolor{gptthreefive}{HTML}{6BAED6}

\DeclareRobustCommand{\colordot}[1]{\begin{tikzpicture}[baseline=(a.south)]
    \node[circle, scale=0.75,color=black, fill=#1] (a) {};
\end{tikzpicture}}

\DeclareRobustCommand{\dashedcircle}{%
    \begin{tikzpicture}[baseline=(a.south)]
        \node[circle, scale=0.75, draw=black, dashed, dash pattern=on 1pt off 1pt, fill=white] (a) {};
    \end{tikzpicture}%
}

\DeclareRobustCommand{\colordotcircum}[1]{\begin{tikzpicture}[baseline=(a.south)]
    \node[circle, scale=0.75,color=black, fill=white] (a) {};
\end{tikzpicture}}

 \title{Causal Strengths and Leaky Beliefs: Interpreting LLM Reasoning via Noisy-OR Causal Bayes Nets}

\author{
 Hanna Dettki \\
 New York University \\
\href{mailto:hmd8142@nyu.edu}{\texttt{hmd8142@nyu.edu}}
}

\date{
  \vspace{-.5em}
  \small WiML Workshop at NeurIPS 2025
}

\begin{document}

\maketitle

\Needspace{10\baselineskip}
\begin{wrapfigure}[28]{l}{0.425\textwidth}
    \centering

    \begin{subfigure}[t]{0.21\textwidth}
        \centering
       
        \includegraphics[width=\textwidth]{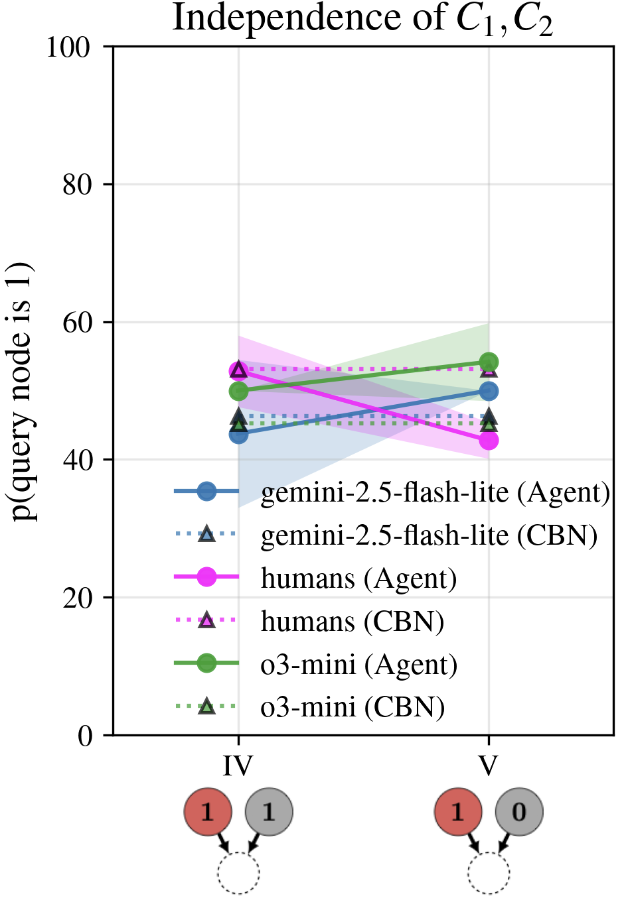}

        \caption{\scriptsize MV \textit{iff} $|IV - V| >0 +\epsilon$}
        \label{fig:comparison_agg_2}
    \end{subfigure}
    \hfill
    \begin{subfigure}[t]{0.21\textwidth}
        \centering
       
        \includegraphics[width=\textwidth]{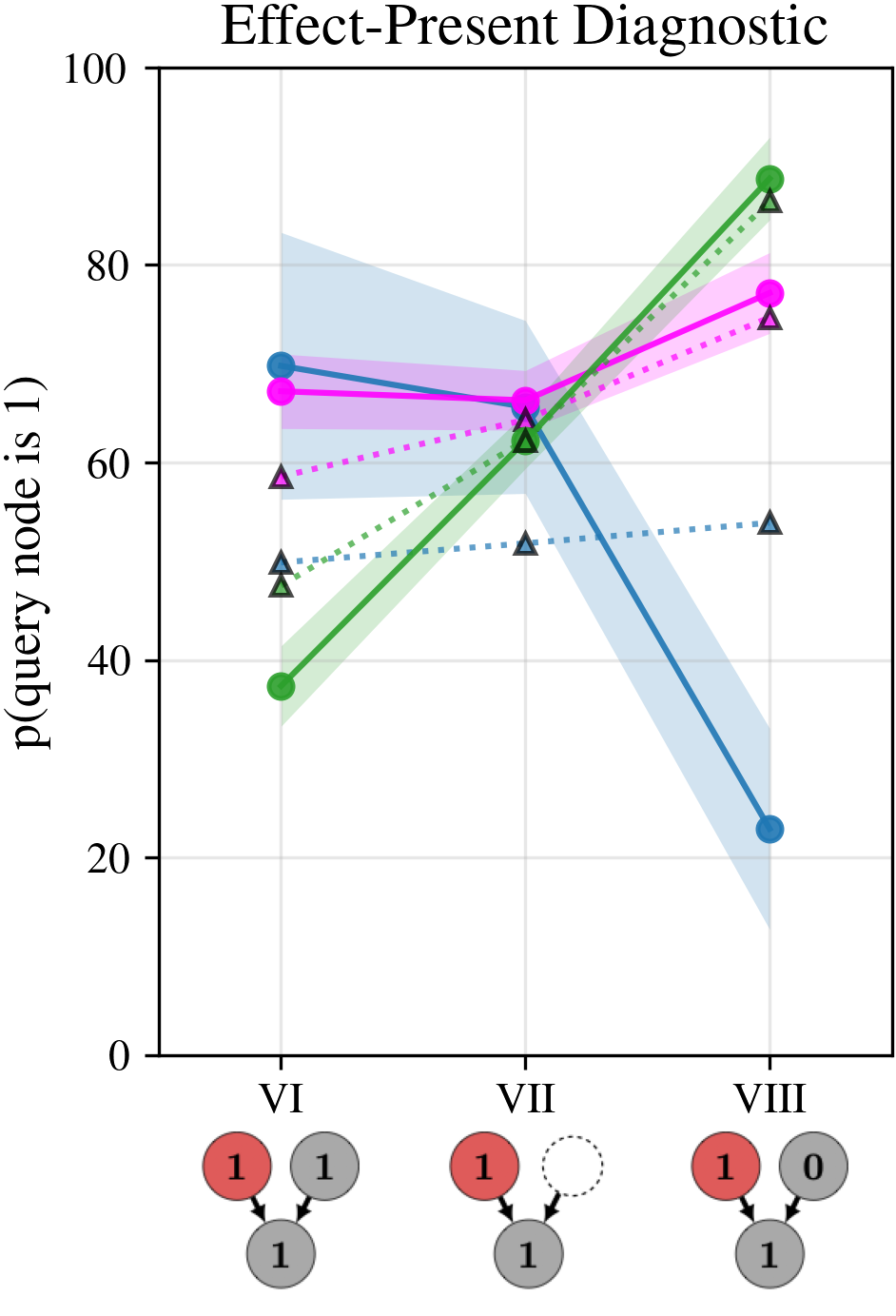}
        \caption{\scriptsize EA iff $VIII > VI$}
        \label{fig:comparison_agg_3}
    \end{subfigure}
    \caption{\scriptsize\textbf{Agents vs. CBN Predictions} Likelihood judgments that query node \colordot{inference} 
    has value 1  $\in \{0,100\}$  of agents' predictions vs. their respective CBN model predictions with  bootstrapped 95\% confidence intervals for agents.
      Graphs on the x-axis visualize a subset of the  conditional probability of the causal inference tasks (I-XI) where the  nodes are colored according to: 
       \colordot{inference} $\to$ query node that the question is asked about; \colordot{observed} $\to$ observed $\in \{0,1\}$; and \dashedcircle{}  
        $\to$  no information on.
        \Cref{fig:comparison_agg_2} shows \emph{Markov violations (MV)} 
       for humans and gemini-2.5-flash-lite, 
       as  $|IV - V| >0 +\epsilon$, visualized by non-horizontal
       lines, where $\epsilon$ is 0.05 in our study. o3 shows no Markov violations and perfect independence of causes.
        \Cref{fig:comparison_agg_3} brings about \emph{explaining away (EA)}, 
       iff $VIII > VI$, visualized by a postitive slope. \emph{o3
       displays perfect EA}, whereas gemini-2.5-flash-lite shows no EA and \emph{humans show weak EA}.
        Experiment: Semantically meaningful (RW17) content, numeric prompt. }
    \label{fig:main_comparison_agg}
\end{wrapfigure}

The nature of intelligence in both humans and machines is a long-standing question. 
While there is no universally accepted definition, the ability to reason causally is often regarded as a pivotal aspect of intelligence \citep{lake2017building}.
Evaluating causal reasoning in LLMs and humans on the same tasks provides hence a more comprehensive understanding of their respective strengths and weaknesses.
\vspace{.25em}
\\ \textbf{Goals, Contributions \& Methods.}
Our study asks: (Q1) Are LLMs aligned with humans given the \emph{same} reasoning tasks (see \citet{dettki2025large,rehder2017failures})? 
(Q2) Do LLMs and humans reason consistently at the task level? 
(Q3) Do they have distinct reasoning signatures?
 We answer these by evaluating 20$+$ LLMs on eleven 
 semantically meaningful causal tasks  formalized by a collider graph ($C_1\!\to\!E\!\leftarrow\!C_2$ )  
under   \emph{Direct} (one-shot number as response = probability judgment of query node     \colordot{inference}   being one (\Cref{fig:main_comparison_agg})) and \emph{Chain of Thought} (CoT; think first, then provide answer).
  Judgments are modeled with a leaky noisy-OR causal Bayes net (CBN) whose \citep{cheng1997causalpower}
     \begin{wrapfigure}[20]{r}{0.42\textwidth}
  \centering
    \centering
    \includegraphics[width=.41\textwidth]{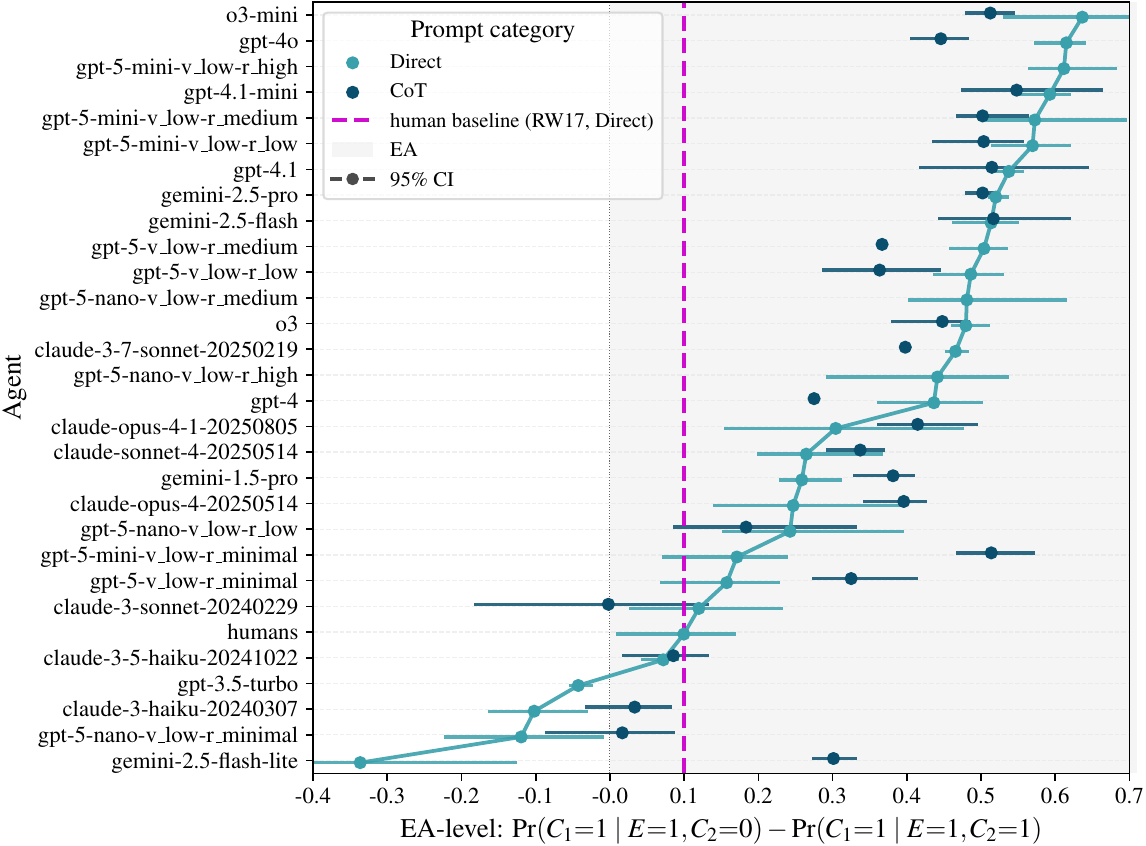}
    \caption{\scriptsize EA levels by agent and semantically meaningful  prompts  both Direct and CoT. EA
  emerges in collider graphs  when evidence for one cause reduces the belief in the other cause, i.e., when $Pr(C_1\!=\!1|E\!=\!\!1, C_2\!=\!1)\!<\!Pr(C_1\!=\!1|E\!=1, C_2\!=\!0)$.
    The naming convention for the GPT-5 familiy is as follows:
    \texttt{gpt-5<suffix>\_v\_<verbosity-level>\_r\_<reasoning-effort>}.}  
   \label{fig:ea_0}
  \end{wrapfigure}
  parameters $\theta=(b,m_1,m_2,p(C)) \in [0,1]$ include a shared prior $p(C)$;
   we select the winning model via AIC between a 3-parameter symmetric causal strength ($m_1{=}m_2$) 
   and 4-parameter asymmetric ($m_1{\neq}m_2$) variant. 
  We explore 3 research questions:
    human-LLM alignment measured by Spearman correlation $\rho$ (Q1), 
   task-level LOOCV-$R^2$ from CBN fits (Q2), 
   parameter-signature profiling $(b,m_1,m_2,p(C))$ (Q3). 
    This separates our work from \Citet{dettki2025large} by replacing the logistic link with leaky noisy-OR, 
    expanding the number of evaluated LLMs ($\sim 5\times$), 
    and enabling clearer evaluations of explaining away (EA) and Markov-violation (MV) diagnostics.  
    EA   emerges in collider graphs  when evidence for one cause reduces the belief in the other cause, 
visually represented as a positive slope in \Cref{fig:comparison_agg_3}.
MV occurs when the presence of one cause affects the belief in another cause, 
violating the independence assumption in a collider structure \Cref{fig:comparison_agg_3}, visually represented by a slope e.g., humans, while \texttt{o3} shows no MV.

  \textbf{Human-LLM alignment: SOTA models establish ceiling; CoT helps others converge (Q1).}
  Recent top-performing LLMs, e.g., \texttt{gemini-2.5-pro}, 
 already show strong human alignment under Direct prompting (\(\rho\approx0.85\)),
  with only little to no improvement via CoT. 
  Conversely, CoT significantly boosts alignment in lighter or older models 
  (e.g., \texttt{gemini-2.5-flash-lite}: \(+0.503 \to \rho=0.845\)), 
   helping them converge to the same ceiling.

\textbf{Humans are consistent reasoners; CoT improves reasoning consistency, especially for smaller \& older models (Q2).}
On our tasks, CoT yields a small but reliable lift in median reasoning consistency, 
raising LOOCV \(R^2\) from \(0.933\) to \(0.941\) \((+\!0.008,\,+0.91\%)\).
More importantly, CoT disproportionately helps the less consistent agents under Direct prompting:
 the lower tail rises (minimum \(R^2\): \(0.277 \text{ (\scriptsize gemini-2.5-flash-lite; numeric)} \rightarrow 0.692 \text{ (\scriptsize claude-3-haiku-20240307; CoT)}\)) and the spread tightens markedly (IQR \(0.116 \rightarrow 0.060\)).
Humans show high consistency across tasks, with LOOCV \(R^2=0.937\) with only a narrow gap to SOTA models who achieve LOOCV$R^2$ of up to .99 (\texttt{gemini-2.5-pro}).

\textbf{Explaining-away is common; CoT effects are mixed (Q3).}
Most LLMs (27/30) show explaining-away ($\mathrm{EA} > 0$), and 24/30 exceed human EA levels
 ($\mathrm{EA}_{\text{human}}{=}0.09$) (see \Cref{fig:ea_0}). 
 CoT helps  agents lacking EA (e.g., \texttt{claude-3-haiku}, \texttt{gemini-2.5-flash-lite}) 
 but can slightly reduce EA in strong ones (e.g., \texttt{gpt-4o}, \texttt{o3-mini}). 
 A similar pattern holds for Markov violations: while eight agents show MV under Direct prompting, CoT improves most but can worsen others
  (e.g., \texttt{claude-3.5-haiku}). High-EA--no-MV agents have low leakage $b$ (0-0.1), strong  causal strength $m_1,m_2$ (0.75-0.99), 
  and midrange priors, while agents with MV or weak EA show higher $b$ (0.15-0.62) and weaker $m_i$ (0.25-0.82).

 \textbf{Outlook.}
Next steps include extending this framework to semantically meaningless tasks and other causal structures beyond colliders to probe reasoning robustness.
It should be noted that ``normative'' parameter regimes (low leak, strong causes) are not universally 
optimal and  ultimately depend on the user-setting: 
tasks that legitimately require uncertainty about unobserved causes may warrant nonzero leak. 
Our prompts do not control this dimension -- we neither instruct models to ignore nor to include 
unmentioned causes.
A targeted analysis of the explanations received through 
CoT could provide first insights into whether and how LLMs represent and regulate them.

\printreferences

\end{document}